\documentclass[sigconf]{acmart}
\AtBeginDocument{%
  }

\setcopyright{acmlicensed}
\copyrightyear{2025}
\acmYear{2025}

\usepackage{xcolor}
\usepackage{lipsum}
\usepackage{tabularx}

\newcommand{\rizwan}[1]{\textcolor{orange}{Rizwan: #1}}

\usepackage{glossaries}
\newacronym{llm}{LLM}{Large Language Model}
\usepackage{booktabs}

\usepackage{listings}
\usepackage{xcolor}

\definecolor{codegreen}{rgb}{0,0.6,0}
\definecolor{codegray}{rgb}{0.5,0.5,0.5}
\definecolor{codepurple}{rgb}{0.58,0,0.82}
\definecolor{backcolour}{rgb}{0.95,0.95,0.92}

\lstdefinestyle{mystyle}{
    backgroundcolor=\color{backcolour},   
    commentstyle=\color{codegreen},
    keywordstyle=\color{magenta},
    numberstyle=\tiny\color{codegray},
    stringstyle=\color{codepurple},
    basicstyle=\ttfamily\footnotesize,
    breakatwhitespace=false,         
    breaklines=true,                 
    captionpos=b,                    
    keepspaces=true,                 
    numbers=left,                    
    numbersep=5pt,                  
    showspaces=false,                
    showstringspaces=false,
    showtabs=false,                  
    tabsize=2
}

\begin{document}

\title{A Cloud-based Multi-Agentic Workflow for Science}

\author{Anurag Acharya}
\affiliation{%
  \institution{Pacific Northwest National Laboratory}
  \city{Richland}
  \state{Washington}
  \country{USA}}
\email{anurag.acharya@pnnl.gov}

\author{Timothy Vega}
\affiliation{%
  \institution{Pacific Northwest National Laboratory}
  \city{Richland}
  \state{Washington}
  \country{USA}}
\email{timothy.vega@pnnl.gov}

\author{Rizwan A. Ashraf}
\affiliation{%
  \institution{Pacific Northwest National Laboratory}
  \city{Richland}
  \state{Washington}
  \country{USA}}
\email{rizwan.ashraf@pnnl.gov}

\author{Anshu Sharma}
\affiliation{%
  \institution{Florida International University}
  \city{Miami}
  \state{Florida}
  \country{USA}}
\email{ashar076@fiu.edu}

\author{Derek Parker}
\affiliation{%
  \institution{Pacific Northwest National Laboratory}
  \city{Richland}
  \state{Washington}
  \country{USA}}
\email{derek.parker@pnnl.gov}

\author{Robert Rallo}
\affiliation{%
  \institution{Pacific Northwest National Laboratory}
  \city{Richland}
  \state{Washington}
  \country{USA}}
\email{robert.rallo@pnnl.gov}

\renewcommand{\shortauthors}{Acharya et al.}
\acmArticleType{Research}

\begin{abstract}
    As Large Language Models (LLMs) become ubiquitous across various scientific domains, their lack of ability to perform complex tasks like running simulations or to make complex decisions limits their utility. LLM-based agents bridge this gap due to their ability to call external resources and tools and thus are now rapidly gaining popularity. However, coming up with a workflow that can balance the models, cloud providers, and external resources is very challenging, making implementing an agentic system more of a hindrance than a help. In this work, we present a domain-agnostic, model-independent workflow for an agentic framework that can act as a scientific assistant while being run entirely on cloud. Built with a supervisor agent marshaling an array of agents with individual capabilities, our framework brings together straightforward tasks like literature review and data analysis with more complex ones like simulation runs. We describe the framework here in full, including a proof-of-concept system we built to accelerate the study of Catalysts, which is highly important in the field of Chemistry and Material Science. We report the cost to operate and use this framework, including the breakdown of the cost by services use. We also evaluate our system on a custom-curated synthetic benchmark and a popular Chemistry benchmark, and also perform expert validation of the system. The results show that our system is able to route the task to the correct agent $90\%$ of the time and successfully complete the assigned task $97.5\%$ of the time for the synthetic tasks and $91\%$ of the time for real-world tasks, while still achieving better or comparable accuracy to most frontier models, showing that this is a viable framework for other scientific domains to replicate. 


\end{abstract}

\maketitle

\section{Introduction}
\label{sec:introduction}
Large Language Models (LLMs) have demonstrated remarkable utility not just in text generation but across an array of complex and niche domains. They have demonstrated capabilities in not just Natural Language Processing (NLP) tasks, but also in complex use-cases like reflection and planning~\cite{park2023generative}, problem-solving~\cite{rasal2024optimal}, and even idea generation~\cite{ege2024chatgpt}. Because of this, they have been increasingly adapted by researchers and professionals across diverse fields including science~\cite{Zhou2023ASOA, Liu2023EvaluatingLLB, Boyko2023AnIOD, Ji2024ACSF, munikoti2024atlantic}. But even as standard LLM use has found massive success and domains are still learning how to utilize their full potential, researchers are quickly discovering that LLM-based agents can offer even more capabilities.

These LLM-based agents, due to their ability to call external resources and tools, are now rapidly gaining popularity, particularly in scientific fields that require expertise beyond what simple LLMs can currently offer. These agents have the ability to execute code, query repositories and databases, crawl the web, and in general interact with most types of computational tools and Machine Learning (ML) modules. As a result, they are already starting to be adapted in various field including general science \cite{schmidgall2025agent}, education \cite{chu2025llm}, data science \cite{hong2024data}, scene simulation for autonomous driving \cite{wei2024editable}, law \cite{li2024legalagentbench}, and cybersecurity \cite{fang2024llm, fang2024llmhack}, among others.

However, with each of these different tools and resources comes a different challenge. When executing code automatically, for example, it is important to ensure that arbitrary code execution cannot escape isolation, access sensitive resources, or be manipulated by nefarious actors \cite{guo2024redcoderiskycodeexecution, he2024securityaiagents}. Similarly, when performing literature review, it has been seen that LLMs often produce plausible but incorrect information due to hallucinations \cite{ravichander2025halogen, tonmoy2024comprehensive, Huang2025}. Overall, the heterogeneity of the various scientific tools and artifacts, the ever-expanding list of models available, and the increasing demand on compute resources means sometimes implementing an LLM-based system in a scientific setting can be more hindrance than help due to the challenge in finding a good workflow. 

In this work, we present a modular, cloud-native workflow for a graph-based multi-agent orchestration to act as a scientific copilot. While the framework we present in itself is domain-agnostic, we present a use case by showing an implementation of it in the catalysis domain. Our framework includes a main agentic framework built using LangGraph and supported by an array of heterogeneous tools, each of which have the ability to perform a specific task that can help scientists in the field of catalysis. The copilot system consists of a central supervisor agent that the human interacts with. The supervisor orchestrates the flow in the system across the six available agents: Literature Review Agent, Segmentation Agent, Simulation Agent, Uncertainty Quantification Agent, Data Analysis Agent and Hypothesis Generation Agent. However, we have built the system to be modular and tool-agnostic, so it can be easily adapted and modified to any desired use case. The tools give the LLM the ability to combine the knowledge and expertise of scientists working in the field. In particular, we demonstrate the power of the LLMs when combined with domain-focused ML models. Our framework combines the know hows of three different ML models developed by different domain-experts working to address problems in catalysis degradation research. For example, one tool is used to track nanoparticles in videos generated from catalysis experiments, whereas another tool is used to predict the size of nanoparticles given some experimental conditions. By concatenating the knowledge obtained from these diverse tools developed by different researchers, the users of the copilot can potentially make broader scientific discoveries. To support the simulation infrastructure, there are also multiple ways for users to feed the experimental data to the system so that it's available for analysis. We explain the entire framework in the subsequent sections. Additionally, we also measure the performance of our framework in terms of the agent invocation and overall task completion and report those results. We recognize that the cost of cloud providers and state-of-the-art LLMs can be a major factor in the people deciding whether or not to implement systems like these. To that end, we report the running cost of the entire setup including the breakdown of the costs by the service. Our hope is that our framework can serve as a blueprint for any future researchers or professionals who want to adapt a similar capability for their domain.

\begin{figure*}
    \centering
    \includegraphics[width=0.8\linewidth]{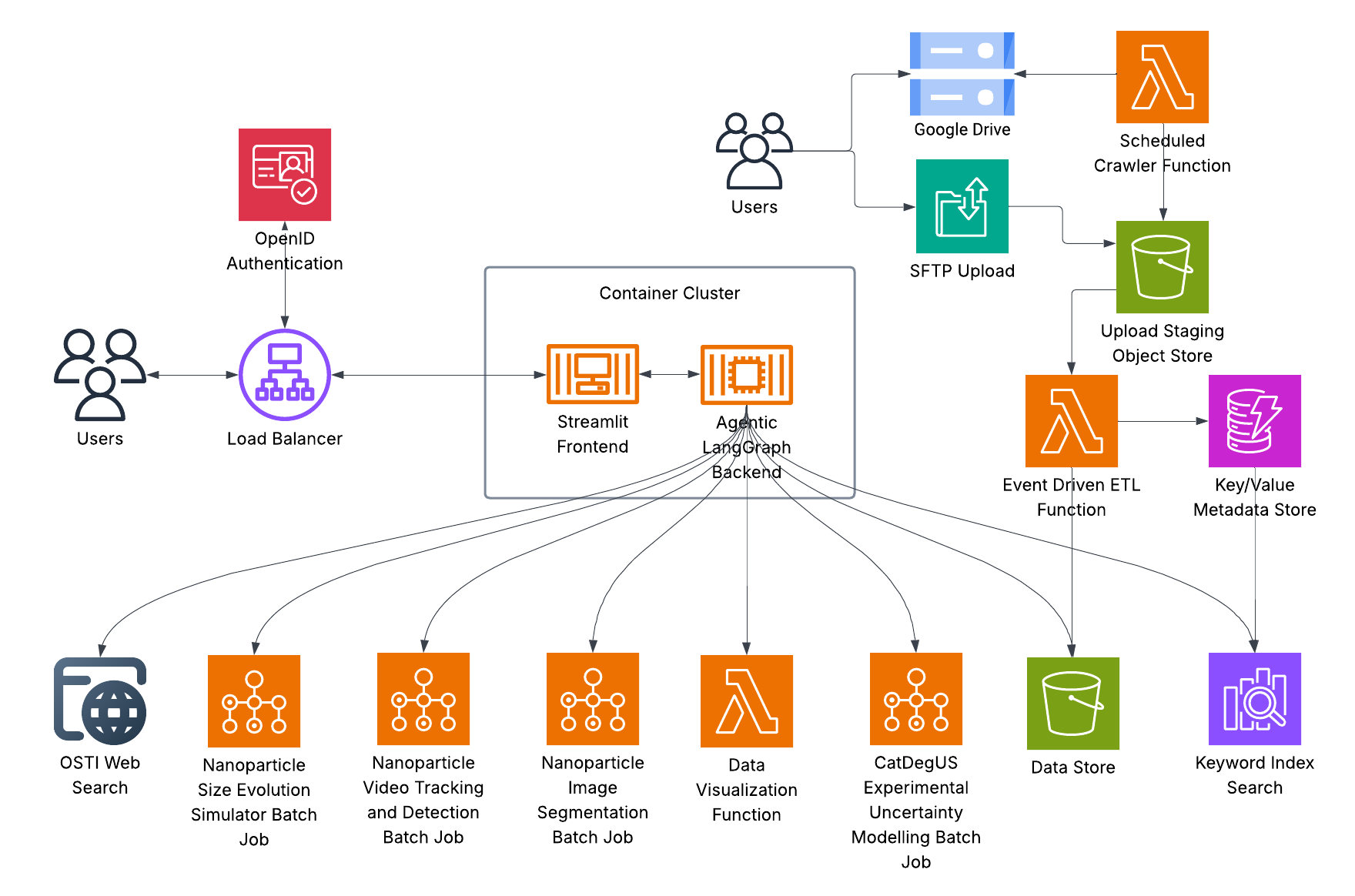}
    \caption{The overall workflow of the entire system. The arrows indicate the flow of data and information.} 
    \label{fig:system-architecture}
    \Description{The overall workflow of the entire system. The arrows indicate the flow of data and information. \rizwan{we need to reference this figure. for UCC, this may be important.}}
\end{figure*}

\section{Related Works}
\label{sec:related}
\subsection{Workflows for AI/LLM for Science}
\label{subsec:related-general}


Workflows \cite{gridach2025AgenticAFA, ren2025TowardsSIA}, a structured sequence of tasks, utilizing LLMs and general AI systems have increasingly enhanced scientific research. \citet{boiko2023autonomous} demonstrates how LLMs can further support chemical research, developing an AI system that is capable of optimizing  experimental designs and execution by incorporating LLMs. Their work demonstrates the potential of utilizing an LLM workflow to reduce manual intervention and streamline decision-making within chemistry related research. Similarly, \citet{bran2024augLLM} explored augmenting LLMs with chemistry specific tools to accomplish tasks across broad chemistry domains such as organic synthesis. \citet{lai2023artificial} integrates LLMs, Bayesian optimization, and an active learning loop to expedite and enhance catalyst optimization and development process. Together, these studies illustrate promising strategies to incorporate LLMs into scientific research. However, this field still lacks both specialized frameworks tailored towards the  study of catalysts and an agentic system that is capable of running solely on the cloud ensuring ease of use for researchers.

\subsection{LLM-based Agentic Workflows for Science}
\label{subsec:related-agentic}


The development of agentic based workflows for scientific research represents a growing yet underdeveloped area of research that can significantly improve scientific advancements \cite{gridach2025AgenticAFA}. \citet{swanson2024TheVLA} presented an example of autonomous agents capable of designing SARS-CoV-2 nanobodies in a "Virtual Lab." This agentic system combined experimental feedback with predictive modeling to iteratively optimize outputs and validate designs. While their work illustrated the feasibility of autonomous workflows for scientific research, its focus on biological systems meant that frameworks for agentic workflows in areas such as catalysis and other domain specific topics are left unexplored and unrepresented.  While pioneering efforts demonstrate potential in specific domains \cite{gao2023retrieval, swanson2024TheVLA, boiko2023autonomous, bran2024augLLM} highlighting the utility of agentic setups, the approaches are narrowly tailored leaving the applicability to other fields unexplored. Despite the advances in agentic frameworks \cite{ren2025TowardsSIA}, the body of work describing comprehensive agentic workflows for supporting  scientific research remains sparse, leaving researchers without robust tools to address challenges in scientific specific domains.
The scarcity of agentic workflows in chemistry related research, and more specifically catalysis research, underscores the need for domain-agnostic innovations. This study seeks to address this gap by designing an agentic framework tailored to further accelerate the study of several areas of science.


\section{System Description}
\label{sec:system}
This section details the architecture and implementation of the system, which is centered on a graph-based multi-agent orchestration layer using the \textit{langgraph-supervisor} library\footnote{\url{https://github.com/langchain-ai/langgraph-supervisor-py}}, which allows hierarchical multi-agent coordination. The design is modular, cloud-native and intended to support execution of core scientific workflows across distributed environment.  

\subsection{Architecture Overview}

The application is hosted on a horizontally scalable, high-availability container runtime which pools instances behind a load balancer. The load balancer distributes requests across the pool in a round-robin fashion. If any of the instances crash, the runtime will provision new instances to maintain a given pool size. The application is made available to end users with a public domain name (DNS) and requires authentication for user requests to traverse the load balancer. DNS is handled by our organization's platforms.

The data ingestion pathway is an event driven pipeline that integrates with the subject matter expert (SME) document sharing platform and minimizes the training burden placed on them. The pathway takes their data and makes it available to the agentic application via tools.

The system employs a graph-based workflow architecture where a central \textit{supervisor} agent delegates tasks to a set of specialized sub-agents. The workflow is executable in two primary modes: (1) \textbf{Full CoPilot}, where user instructions are routed through an orchestrating agent which delegates tasks to downstream agents, and (2) \textbf{Direct Tool Mode}, which allows the user to invoke a single tool that does only one specific scientific task. This dual-mode operation supports both exploratory usage and quick execution of established workflows.

The agents currently implemented in our framework 
are discussed in detail in \S\ref{sec:agents}. Additional agents can be registered by extending the graph configuration. Intermediate results are shared across agents via a memory mechanism, and task transitions are governed by tool outputs and LLM completions.

\subsection{Infrastructure}
\subsubsection{Compute}
The software artifacts that are part of the pipeline are containerized and hosted on serverless compute platforms. In each part of our application, compute is selected based on the component's expected life-cycle. Short-lived jobs are hosted on function platforms. Long-running jobs are hosted on batch platforms. Services that handle user requests are hosted on scalable, high-availability compute platforms and are pooled behind load balancers as shown in Figure~\ref{fig:system-architecture}.

\subsubsection{User Access}

A lot of experimental data produced by the SMEs is proprietary. Since this data is used to derive machine learning models utilized in the agentic framework. A requirement was to manage user access to the web front-end of the framework. As a side effect, it also helped us to manage costs associated with using the cloud infrastructure. 

For our user services, we delegate authentication (authN) to our organization's authN service. The organizational authN is then integrated into our load balancer to ensure that all users who traverse the load balancer are authenticated. If not, then the load balancer redirects unauthenticated users to a login page. We did not discover a need for fine-grained authorization (authZ). Instead of granting users permissions for specific features in the application, users are granted permission for the entirety of the application.

We have two ingestion pathways. The first one is a Google Drive upload, in which case Google handles its own user authN/Z. For our SFTP upload, we setup PKI for users on a case-by-case basis.

\subsection{Application}
\subsubsection{Components}

The application front-end is a Streamlit application\footnote{\url{https://streamlit.io/}} that uses a Streamlit chatbot component standard library. The back-end is a LangGraph\footnote{\url{https://langchain-ai.github.io/langgraph/\#langgraphs-ecosystem}} application. The back-end and front-end are hosted on the container cluster as shown in Figure~\ref{fig:system-architecture}. Originally, the application was deployed as a single monolithic process. But since we needed to expose an API to run an automated evaluation tool against the agentic system, the application was factored into a front-end and back-end components and a FastAPI layer\footnote{\url{https://fastapi.tiangolo.com/}} was added to the back-end.

The back-end makes itself discoverable to the network via an internal load balancer not pictured in Figure~\ref{fig:system-architecture}. The front-end could integrate with the back-end via a container cluster internal network, but instead both the evaluation platform and the front-end integrate with the load-balancer endpoint. This is to ensure that the evaluation platform integrates against a point that is frequently exercised and is more likely to be stable.

\subsubsection{Foundation Models}

Models used by the application are hosted on AWS (Claude 3.7 Sonnet\footnote{\url{https://www.anthropic.com/news/claude-3-7-sonnet}}) and Azure (OpenAI o3-mini\footnote{\url{https://openai.com/index/openai-o3-mini/}}). Agents requiring code generation and code execution capabilities have been paired with Claude, since prior research has shown that Claude models perform better than most major LLMs on code generation \cite{jiang2024survey}, while all the others use o3-mini for text-based reasoning tasks. 

\subsection{Ingestion}

\subsubsection{User Upload}
We've collaborated with our SMEs to define a data package schema which we are able to parse and ingest into our system. We have a user-interface tool where users can fill out a form and upload data files to create the aforementioned package. 
Users can then upload the package either directly into our system via secure file transfer protocol (SFTP) into an object storage folder or they can continue to use their regular document sharing platform which we crawl at a fixed schedule.

Using the above streamlined process, we are able to ingest all the information required for the development of machine learning models and understand the various catalyst degradation mechanisms that might be observed during experiments. The metadata schema was designed with SMEs to accommodate a broad range of catalysis experiments and different data types that might be produced during these experiments. We are able to capture both experimental and computational characterizations as well as the catalyst degradation mechanisms that are observed.
For example, we accommodate both DRIFTS and XAS data from spectroscopic techniques used to characterize catalysts and understand reaction mechanisms. Generally, this method allows for standard data ingestion such that this information can be made available to the users in a meaningful way and linked to the agentic framework.

The metadata schema is defined in Python using the Pydantic \cite{pydantic} data validation library. It allows us to create a typesafe definition of our metadata. When a developer references the Pydantic schema, they are able to see all of the fields that may be populated at-a-glance and they can save time which might otherwise be spent reading through code that incrementally constructs a Python dictionary object. Our model is configured to allow null entries for all fields to tolerate inconsistency in schemas and match the expected flexibility of working with raw JSON or native Python dictionaries. Pydantic also natively supports serialization and deserialization between its models and JSON.


\subsubsection{Extract Transform Load (ETL)}
We have an ETL function that decomposes the data package back into metadata and experimental/computational data. The metadata is stored in a NoSQL key/value data store and the experimental/computational data is stored in an object store. An event driven pipeline is setup to automatically index the key/value metadata into a keyword search index. At this point the data is ready to be served to the agentic application.

\subsection{Agentic Architecture}
\label{sec:agents}

\begin{figure}
    \centering
    \includegraphics[width=0.95\linewidth]{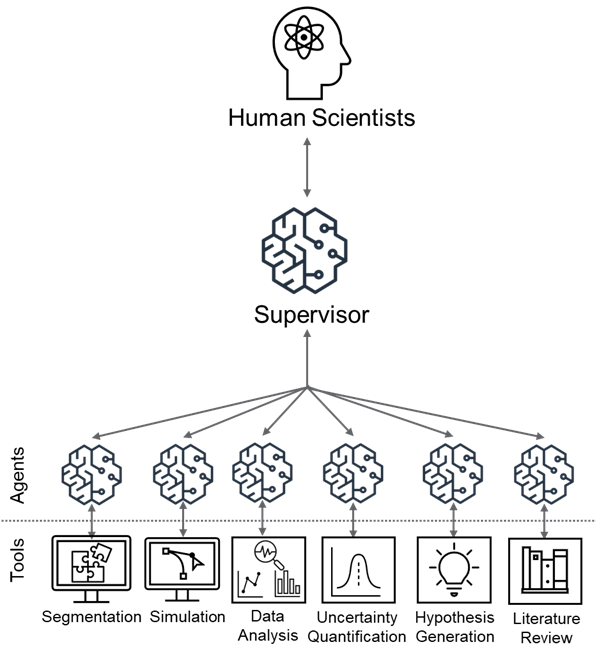}
    \caption{A simplified architecture of the agentic framework of the system from an end-user perspective.}
    \label{fig:copilot-arch}
    \Description{A simplified architecture of the agentic framework of the system from an end-user perspective.}
\end{figure}

The system incorporates multiple agents, each encapsulating a discrete scientific function. All agents are stateless and are invoked as a part of a structured execution graph managed by the supervisor. Each agent has access to tools which can be used to call some external service or an LLM. Intermediate outputs are passed between agents via shared checkpointer memory. Agents are implemented in a ReAct \cite{yao2023react} style format -- receiving task prompts, invoking one or more associated tools, and returning structured outputs. This architecture as shown in Figure~\ref{fig:copilot-arch}, allows each agent to operate as a single-responsibility unit, with execution controlled by the supervisor. 

\subsubsection{Supervisor}

The Supervisor serves as the central orchestrator in our multi-agentic workflow system. It is implemented using the \textit{langgraph-supervisor} library by LangChain, which enables control over agent execution, graph transitions, memory, and prompt routing. The agents are stateless but graph runtime state is preserved to allow human-in-the-loop capability. Unlike systems with fixed routing policies, our supervisor interprets user inputs in context using LLM and delegates them to the most appropriate sub-agent.

\subsubsection{Literature Review}

The literature review agent, \textit{Researcher}, is designed to assist with scientific literature review and summarization. In order to avoid problems with hallucinations, it does not simply rely on LLM's capabilities to perform literature review. But rather it leverages a custom tool that queries the U.S. Department of Energy's OSTI\footnote{\url{https://www.osti.gov/}} (Office of Scientific and Technical Information) repository via a public API.

Upon receiving a user query, the agent invokes the tool, which constructs a request to the OSTI API endpoint and returns a set of publication records along with their metadata. Each record includes title, description, author information, and DOI among other information. The tool then formats and sends the results to the agent which then uses the LLM to summarize, refine, and organize the results before forwarding it to the supervisor.  

\subsubsection{Data Analysis}

The data analysis agent, \textit{Analyzer}, is responsible for the task of exploratory data analysis. The workflow begins outside the copilot: collaborators upload their datasets to a shared cloud storage service as described earlier. 
Simultaneously, the metadata associated with each file (e.g., experiment description, title, etc.) is extracted and indexed in a searchable metadata store with `Keyword Index Search' tool shown in Figure~\ref{fig:system-architecture}.

When a user initiates a query via the chat interface, the agent first performs a metadata lookup against the store to identify relevant files. Once a match is found, the associated file is retrieved from the internal storage bucket for further processing. The Analyzer agent then delegates the `Data Visualization' tool (see Figure~\ref{fig:system-architecture}) to inspect the file to generate an analysis. This includes both natural language analysis and recommendations, as well as Python code that performs relevant statistical analysis and plot generation. The generated code is passed to a sandbox execution environment, which evaluates the code and returns any generated figures. The figures are also stored in object storage for persistence. The results are then compiled and sent back to the supervisor for downstream integration.

To address concerns about the safe execution of LLM-generated code, we implemented a two-tier filtering mechanism. First, we leverage the built-in guardrails provided by our LLM deployment platform to automatically block input and output containing sensitive operations or keywords such as \lstinline!eval, exec, open(, input(, subprocess!, as well as common personally identifiable information (PII) and attack vectors such as access keys or IP addresses. These guardrails prevent potentially unsafe or privacy-violating code from ever reaching the copilot in the first place. Second, after the LLM generates the code and before any execution occurs in the sandbox, we conduct an additional scan on the code. This post-generation filter strips all the imports from the code and blocks execution if any sensitive terms like \lstinline!os, boto3, __import__! are detected. Only a minimal safe set of libraries (\lstinline!numpy, pandas, matplotlib, seaborn!) are permitted. This layered approach ensures a robust defense against the risk of arbitrary code execution and data leakage.

\subsubsection{Hypothesis Generation}

The hypothesis generation agent, \textit{Hypothesizer}, is responsible for formulating structured scientific hypotheses and the corresponding research plans. The agent uses a dedicated LLM tool that returns a research plan, including objectives, theoretical framing, and proposed hypothesis based on the user's query.

\subsubsection{Simulation}
\label{sec:simagent}

The simulation agent is responsible for helping users understand catalyst nanoparticle sintering behavior using hybrid Bayesian modeling techniques that combine physics-based kinetics with experimental data insights. The model can predict nanoparticle size evolution over time in catalyst systems at the specified temperature. 

The modeling framework that includes the training of machine learning model is written in Julia programming language. For the purposes of the agent, inference is performed to generate time evolution of nanoparticle sizes at user-specified temperature. The inference is performed in the compute infrastructure in the cloud using the `Nanoparticle Size Evolution Simulation Batch Job' tool shown in Figure~\ref{fig:system-architecture}. The scalability of the cloud compute batch infrastructure allows users to submit multiple jobs in parallel without burdening the agentic framework. A Docker image is pulled from a container registry and deployed on the compute node seamlessly. The agentic framework uses the tool to trigger the simulations in the cloud infrastructure. The user-specified temperature input is passed as an argument to the job submission request.

The agentic framework also has a tool at its disposal to query the status of the job once it has been submitted. Due to the memory architecture of the framework, a single user can submit and track multiple jobs. Once jobs have finished processing, the user can use a tool to query the results of the simulation. In this case, the compute job writes its outputs to a storage location and available tools gives the ability to locate these outputs. Both text and figures are available. The text output which in this case is raw data, e.g., nanoparticle size with respect to time including 95\% confidence interval, is returned as output of a tool for interpretation by the LLM agent. This is where the agent is able to use its domain knowledge of catalysis to present the results of the simulation output in a meaningful way. Users have the ability to provide follow-up prompts to the agent or perform comparison across multiple simulations performed at different temperatures. In case the user wants to visualize the results, the LLM can provide downloadable links to artifacts produced as a result of the simulation that are stored in an object storage. 

\subsubsection{Segmentation}
\label{sec:segmentationAgent}

The segmentation agent is responsible for automated analysis of nanoparticle behavior in images and videos produced from catalysis experiments. At the heart of this analysis pipeline is the Segment Anything foundation Model (SAM)~\cite{ravi2024sam2}, that provides the ability to do prompt-based visual segmentation. In case of videos, the segments can be tracked throughout multiple video frames using object memory in SAM. The initial prompts in the first frame are provided to SAM using YOLO's~\cite{yolo11_ultralytics} image segmentation model. Once segments or nanoparticles are identified, particle sizes and centroids can be tracked throughout all video frames. Temporal changes in size evolution are available after the completion of the video analysis. An annotated video file is also produced that can be later analyzed by the user.

The image analysis includes the ability to report different shape descriptors of the nanoparticles like, eccentricity, sphericity, and solidity that can be interpreted by the LLM agent in the context of catalyst performance. Both the image and video analysis pipelines are written in Python and require GPUs to run the compute-intensive SAM model. In our framework, these compute-intensive jobs are executed using the `Nanoparticle Size Evolution' and `Nanoparticle Video Tracking and Detection' Batch Job tools as shown in Figure~\ref{fig:system-architecture}. The infrastructure is setup to pull a Docker image from a container registry that contains both the image and video analysis pipelines. The agentic framework uses the tools available to it to submit the analyses based on user requests. In this case, the input images and videos available for analyses are uploaded to a object storage that the job has access to. The agentic framework has a tool available to it so it can list all available inputs to the users along with their experimental conditions. 

In a similar manner to the simulation agent, the segmentation agent has access to tools that can list the status of jobs, and access outputs produced from the analyses. 
This includes both text output that can be presented in the context of catalysis science and annotated images/videos produced as a result of the analysis.

\subsubsection{Uncertainty Quantification}

This agent has a set of tools available to it that facilitates Gaussian Process-based uncertainty analysis as shown in Figure~\ref{fig:system-architecture}. The analysis is conducted using preprocessed time-on-stream catalyst testing data that is used as the training dataset. The user can specify target performance metrics and experimental parameter ranges such as temperature bounds, metal loading specifications and synthesis method. As a result of this analysis, the most informative experimental conditions are produced that lead to maximum learning efficiency and reduced experimental costs. 

The computational framework is setup in a manner similar to those for simulation (\S\ref{sec:simagent}) and segmentation (\S\ref{sec:segmentationAgent}) agents. A docker image is pulled from a container repository whenever the agent wants to perform an analysis. The batch job submission system is able to deploy the compute job using the image. Once again, the scalability of the compute infrastructure allows us to serve multiple users seamlessly. 

The testing data is made available to the modeling framework through object storage that the computational analysis job has access to. Upon completion of the job, the produced outputs are available to the agent in both text and figure formats. The ranked experimental suggestions and 3D uncertainty maps to guide efficient catalyst testing are available for the agent to interpret and present to the user. The prompts to the agent indicate it to provide comprehensive interpretation of recommended experiments in the context of their catalyst development objectives. Ideally, the agent is able to help users actionable experimental guidance with resource efficiency considerations. 

\section{Evaluation}
\label{sec:evaluation}
In this section we present the associated costs and the evaluation results of our system.

\subsection{Cost Metrics}

\begin{figure*}
    \centering
    \includegraphics[width=0.8\linewidth]{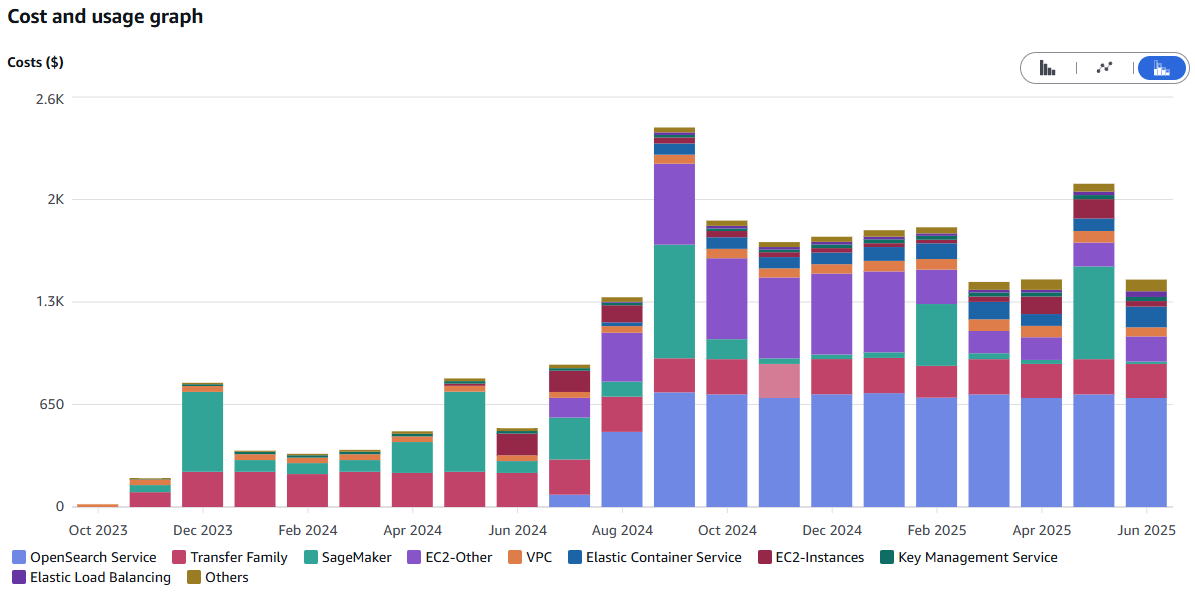}
    \caption{Total monthly costs of the system. The cost to support 100 users is estimated to be 1,750 USD $\pm$ 250 USD.}
    \label{fig:costs}
    \Description{Total monthly costs of the system}
\end{figure*}

Initial costs ramped up over time until we reached a steady-state monthly cost which hovered around 1,500 USD. Before August 2024 we were building a prototype and didn't have a user-facing application. There were common fluctuations in price around compute resources provisioned on an ad-hoc basis for development. Steady-state costs were largely dominated by the keyword index which tended to account for roughly fifty-percent of the monthly costs even in a minimal serverless configuration. Costs could have further been reduced by migrating all users to use the Google Drive upload ingestion method and removing the SFTP upload method which accounted for roughly 200 USD monthly. See Figure~\ref{fig:costs} for a more detailed breakdown of the costs.

\subsection{Agent Invocation Tests}
Our agentic framework has evolved over time to include new simulation models. As the framework scaled up, we needed to test whether the LLM agent was able to correctly invoke the newly integrated tools. For this purpose, we built a synthetic evaluation dataset consisting of test queries to provide to the system to evaluate whether or not the supervisor was invoking the correct agent and its corresponding tools. We are not testing whether the copilot correctly completed the task, but rather if the copilot was able to route the query to the correct agent capable of solving it. 

\subsubsection{Test Dataset Creation}

This evaluation dataset serves as our primary benchmark for assessing whether the copilot correctly invokes the appropriate agent and set of tools for a given task. We generated 20 cases each for the six implemented agents, 120 cases in total, using OpenAI's o4-mini\footnote{\url{https://openai.com/index/introducing-o3-and-o4-mini/}} model and Anthropic's Claude Sonnet 4\footnote{\url{https://www.anthropic.com/news/claude-4}}. These models were chosen to ensure test dataset was not biased toward the same foundation models powering the deployed copilot, while leveraging their strong instruction following and reasoning capabilities. To ensure alignment with agent capabilities, we reused the same prompts used to implement each tool-using agent, modifying only the agent-specific portion. An example of the prompt used to generate the cases is shown in Listing~\ref{lst:test-gen}. 

\begin{lstlisting}[label=lst:test-gen, language=Python, caption=Prompt provided to the LLMs to generate the test queries for the literature review agent.]
prompt = """I want to focus on basic functionality tests for now of a research assistant for catalysis science. I need to test tool invocations. the supervisor agent calls other agents that perform specific tasks. For the performance of the tasks, various tools are available. Give me 20 distinct cases to test the agent. 
Here are the prompts for the agents I will be testing:
* Literature Review Agent
You are a researcher whose job is to perform literature review using available search_osti tools and provide information alongside citation. Use this tool to query the OSTI repository for documents related to a topic."""
\end{lstlisting}


We found that the standalone LLMs were able to generate adequate test prompts to evaluate the correct integration of the newly developed agents and tools. An example sequence of test prompts for the literature review agent is listed in Listing~\ref{lst:test-example}.

\begin{lstlisting}[label=lst:test-example, language=Python, caption=LLM-generated test prompts for the literature review agent.]
# Simple keyword search
"Find recent articles on TiO2-supported Pt catalysts for CO oxidation and give me titles, authors, year, and a brief summary with citations."

# Synonym/abbreviation handling
"Look up catalysts for the water-gas shift reaction (WGS) and report the top five most-cited articles."

# Chemical-formula query
"Search OSTI for papers on NiFe layered double hydroxide catalysts, include at least one citation in standard format."
\end{lstlisting}

\subsubsection{Results}

\begin{table}[t]
\centering
\caption{Agent invocation test performance across agents} 
\label{table:agent_invocation}
\begin{tabular}{llllll}
\toprule
\bf Agent                  & \bf Task Successful    & \bf Correct Agent  \\
\midrule
Data Analysis              & 18                     & 18          \\  
Hypothesis Generation      & 20                     & 19          \\  
Literature Review          & 20                     & 20          \\ 
Simulation                 & 19                     & 15           \\ 
Segmentation               & 20                     & 18           \\ 
Uncertainty Quantification & 20                     & 18          \\  
\midrule
\textbf{Total}          & 97.5\%                   & 90\% \\                                     
\bottomrule
\end{tabular}
\vspace{-1em}
\end{table}

The test cases were executed programatically through the copilot's API rather than the user interface to ensure consistency and to automate test case logging. For each test prompt case, an addendum prompt was appended to instruct the system to return the name of the agent(s) and tool(s) used during the task execution. This allowed us to verify whether the supervisor agent correctly routed the query to the appropriate sub-agents.

Table~\ref{table:agent_invocation} summarizes the performance of the copilot in routing and executing the 20 test cases per agent. Overall, the system achieved a high average task completion rate of 97.5\%. A task was marked as successful if the agent produced a non-empty, non-refusal response within the 10-minute timeout window. This relatively long timeout was necessary to accommodate agents executing computationally extensive tasks such as simulation and analysis.

Correct agent routing--whether the intended agent for that task was invoked--was achieved in 90\% of all cases. Literature Review agent achieved perfect routing accuracy. Data Analysis, Hypothesis Generation, Segmentation and Uncertainty Quantification had greater than equal to 90\% accuracy. In contrast, the Simulation agent had the lowest routing performance, with correct agent invocation rate of 75\%. A detailed analysis of the misrouted cases revealed that these errors were primarily due to ambiguity in task prompts. For example, query such as \textit{"Give me statistical analysis of particle sizes and their changes over time from my tracking results."} was routed to the Data Analysis agent based on the term ``analysis", even though the correct agent should have been Segmentation which handles nanoparticle behavior jobs. This indicates a need for more context-aware routing mechanisms and reflects limitations in the current supervisor agent's routing prompt. 

All the cases where the framework was unable to respond successfully to the task assigned were due to timeout, i.e., 3 cases out of 120 cases. We noted two cases of hallucination: once for a prompt meant for Simulation agent and another one meant for Hypothesis Generation agent. In a couple of prompts meant for Simulation agent, we noticed that the Supervisor agent was not able to direct the instructions to any agent. In a few cases for Simulation, Segmentation, and Uncertainty Quantification agents, the wrong agent was invoked by the supervisor. This is due to the fact that all these agents use the batch job submission system and using the "job" keyword in the prompt likely causes the Supervisor to incorrectly route the tasks.

It is important to note that this evaluation was conducted in a static setting without human intervention. In practice, the deployed system includes a chat-based, human-in-the-loop interface that allows users to clarify prompts or route misclassified tasks, and we expect routing accuracy and task success rates to further improve under realistic user conditions.

\subsection{Domain Evaluation}
\label{subsec:domain-eval}
To evaluate how effectively and successfully our system can perform a domain-relevant task, we ran our full copilot on the ChemBench~\cite{mirza2024large} benchmark -- a suite of chemistry-related queries spanning multiple sub-fields like analytical chemistry, organic chemistry, material science, etc. In addition to benchmarking LLM accuracy and correctness in the given task, an additional focus in this evaluation was on task completion, i.e., whether the copilot was able to perform the task requested of it and give the desired output, regardless of whether or not the answer was factually correct, as this would help us evaluate the usefulness of the copilot infrastructure itself.

Across 2786 benchmark queries, the copilot completed 2521 responses, which is a success rate of $90.49\%$. A completed response is the number of times the question prompted was successfully answered, whether correct or incorrect.\footnote{There were five cases ($0.18\%$) of errors in Toxicity and Safety category occurring on the LLM provider end due to Azure OpenAI's content management policy.} The overall correctness of the system was at $61\%$, which is better than or comparable to every single frontier models reported in the original paper, with the best reported model OpenAI's o1 only being marginally better at $64\%$. A breakdown of both the success and correctness among the various subtopics within ChemBench is shown in Table \ref{table:copilot_success}, along with its ranking when compared to the performance of the 31 different models with performance reported on ChemBench, and the delta from the best-performing model. We see that copilot performs better than most frontier models and is always close to the best model across all sub-categories.\footnote{For the full results on all 31 models, please see the original ChemBench paper~\cite{mirza2024large}.}

\begin{table}
\centering
\caption{Success and correctness rate of the copilot across ChemBench's subtopics. The delta is from the best-performing model, and the rank is based on correctness.}
\label{table:copilot_success}
\begin{tabular}{lc|ccc}
\toprule
\bf ChemBench Topics     & \bf Success & \bf Correct & \bf $\Delta$ & \bf Rank \\
\midrule
Analytical Chemistry & 82.2\%  & 59\% & -3\% & 2 \\
Chemical Preference  & \textbf{93.1\%} & 56\% & -4\% & 5 \\
General Chemistry    & 87.2\%  & \textbf{93\%} & +0\%  & \textbf{1} \\
Inorganic Chemistry  & 91.3\%  & 84\% & -6\% & 2 \\
Materials Science    & 90.5\%  & 74\% & -6\% & 2 \\
Organic Chemistry    & 92.1\%  & 85\% & \textbf{+3\%} & \textbf{1} \\
Physical Chemistry   & 84.2\%  & 82\% & -7\% & 2 \\
Technical Chemistry  & 82.5\%  & 75\% & -10\% & 3 \\
Toxicity and Safety  & 90.1\% & 42\% & -6\% & 3 \\
\midrule
Overall              & 90.5\% & 61\% & -3\% & 4 \\
\bottomrule
\end{tabular}
\newline
\vspace{-1.75em}
\end{table}


\subsection{Subject Matter Expert Validation}

To complement benchmark-based evaluations and to test our system on realistic scenarios, we validated our system by having 
four domain experts use our copilot in a standardized setting and qualitatively examining the output. These experts were at various level of their career, from early career scientists with 2-3 years to experience to more established experts with 20+ years of experience.

We asked each expert to create a case to test the copilot that would represent a task they would normally undertake manually in their daily research work. To standardize this process, we asked the experts to write down the case using a structured case template. This included a detailed task description, a clear explanation of how the expert would approach the problem if done without copilot, a definition of what constitutes a successful outcome for the task, and a difficulty rating labeled as Easy, Medium, or Hard.

The experts reported that the system is able to complete the tasks in all but one case. In the failed case, the copilot was not able to route the task to the hypothesis generator agent successfully. But when the user manually selected the agent directly, it was able to produce the desired result.

\section{Discussion}
\label{sec:discussion}


In this section, we discuss some of the challenges and issues we encountered, as well as the reasoning for some of the decision choices we made when designing and developing this system.

\textbf{Adjusting ingestion to meet user requirements}: It was difficult to drive user adoption of the SFTP upload process. It is likely that we would have ingested user data into our system sooner and more frequently if we had started with the Google Drive crawler approach since it doesn't burden a non-technical user with new process. At the beginning of the project there wasn't yet an attractive feature that would have motivated a user to undertake the learning effort.

\textbf{Splitting application:} The application was originally a monolith that included both front-end server-side rendering code and back-end agentic LangGraph code. Splitting the application into front-end and back-end components was a refactoring we performed before exposing an API from the back-end that our evaluation platform could integrate against. It's not strictly necessary to split and its possible to expose an API from a monolithic application, but splitting preserves a clearer engineering pathway towards horizontally scaling the back-end. Another reason is to ensure greater stability of the API that is exposed from the back-end. In the monolithic configuration, the API is only ever exercised by batch jobs that are occasionally run on the evaluation platform. In the split configuration, the API is also exercised by the front-end which means user and developer traffic implicitly exercise the API.

Keeping the application whole as a monolith makes sense for some teams. The split means more cloud infrastructure integration failure points, so if there is a lack of available staff to debug cloud infrastructure integration issues, then it could be beneficial to keep the application whole.

\textbf{LLM-generated prompts for agents}: The individual agents that are responsible for separate tasks need elaborate prompts for them to function as intended. If the prompt is properly designed to provide all the information and resources available to the agent, the user queries can be served adequately and tasks can be routed correctly to the right set of tools. To overcome some of the challenges associated with providing prompts to the agents, we utilized LLM (Claude Sonnet 4 model) to generate prompts. The prompts were fine-tuned to include limitations of the tools. For example, the prompt utilized for the segmentation agent is listed in Listing~\ref{lst:prompt}.

\begin{lstlisting}[label=lst:prompt, language=Python, caption=Segmentation agent prompt]
segmentation_agent = create_react_agent(
    llm,
    tools=[submit_batch_job_videoTracker, submit_batch_job_imageSegmenter, get_batch_job_status_nonblocking, get_batch_job_list, check_s3_output_imageSegmentation, check_s3_output_imageSegmentation_links,  check_s3_output_videoTracker, check_s3_output_videoTracker_links, list_s3_files],
    name = "segmenter",
    prompt="""You are an expert computer vision and catalysis research assistant specializing in automated analysis of nanoparticle behavior in experimental images and videos.
    Your role is to help researchers extract quantitative insights from visual catalysis data using advanced image processing and particle tracking techniques.
    ## Your Expertise: ...
    ## Available Tools: ...
    ## Your Responsibilities: 
        ### File Selection and Management: ...
        ### Image Analysis: ...
        ### Video Analysis and Tracking: ...
        ### Data Interpretation: ...
    ## Communication Style: ...
    ## Key Guidelines: ...
    ## Tool-Specific Capabilities: 
     **Video Tracking**: Automatically detects all nanoparticles in the first frame, then tracks their sizes and centroid positions throughout the video sequence.
     Produces both quantitative datasets and an annotated video showing visually marked particles for verification and presentation purposes.
     
     **Image Segmentation**: Provides detailed morphological characterization of individual particles including area and shape descriptors for static image analysis.
     
     When users request analysis, first help them browse and select appropriate files from S3 storage, then guide them through the analysis process, explain what insights can be extracted, and help them understand the implications for their catalysis research.""")
\end{lstlisting}

\textbf{Structured input for tools}: To provide structured inputs to the tools, we utilized Pydantic's \cite{pydantic} \textit{arg\_schema}. This helped us define expected input arguments for our tools since in most cases the agent needs to call the tool with specific inputs. Using the definition, we provided default values and elaborate description of the input arguments for the LLM agent to process. For instance, we could limit the temperature input to have specific units (e.g., Celcius) as expected by the model. In this way, even if the user provided a different unit to describe their input, the agent could convert the units and provide the proper arguments to the tool. An example of the input for video tracking analysis is listed in Listing~\ref{lst:args}. The description of the tool can also be provided through the docstring. The LLM can parse both the docstring and the args\_schema as desired. Both serve to clarify the purpose of the tool and its various arguments, so that it can used adequately by the LLM agent. 

\begin{lstlisting}[label=lst:args, language=Python, caption=Use of Pydantic args schema for description of the argument of the video tracking tool.]
from langchain_core.tools import tool
from pydantic import BaseModel, Field

class VideoTrackerInput(BaseModel):
    input_video: str = Field(description="Video file from catalysis experiment to be processed for automatic tracking of nanoparticle movement and behavior throughout the recorded sequence")

@tool(args_schema=VideoTrackerInput)
def submit_videoTracker_job(input_video: str) -> str:
"""
    This tool submits nanoparticle tracking analysis as a batch job to AWS cloud infrastructure for automated video processing. ...
"""

\end{lstlisting}

\textbf{Advantages of using the cloud infrastructure}: Having compute-intensive jobs in our agentic framework deployed in the cloud give us the ability to serve multiple users across disparate locations. The scalability of cloud infrastructure gives us the ability to provision compute resources as required to execute the simulations. We are able to run different types of simulations in the infrastructure by using containers. In the setup, we containerize an application and store the image in a container repository. For each simulation that needs a different image, we establish a job definition that links to the image and scheduling policy for the underlying compute resources. Once this setup has been deployed, we can utilize tools in the copilot to deploy batch jobs in the cloud. In this way, our framework is able to support computing requirements of various kinds of simulations. For example, in this work, the simulation agent is able to deploy CPU-based Julia simulations and the segmentation agent is able to deploy GPU-based Python simulations. 

The I/O requirements for our simulations are satisfied by using object storage that the container applications can access while running on the cloud. In this way, the data needed and produced by our simulations remains securely on the cloud resources. Users are unable to access this data directly, unless web-links are provided to the files that maybe downloaded on request. This model keeps the compute environment secure and users isolated from the underlying complexities of the infrastructure. At the same time, we are able to support running a variety of jobs in the background and the time to adopt a new simulation pipeline is relatively quick. 

\textbf{Cloud-agnostic setup}: While our system is currently deployed on AWS infrastructure for compute, it remains cloud-portable by design. All services and equivalent components such as storage can be swapped with alternatives from other providers or on-premise solutions without significant engineering overhead. This architectural choice allows users with varying infrastructure capabilities to adopt the system with minimal friction.

\section{Conclusions and Future Work}
\label{sec:limitations}
In this work, we demonstrate an end-to-end multi-agentic framework successfully deployed in the cloud environment that is able to assist catalysis researchers across multiple institutions. We combined the expertise of researchers through different machine learning models and computational tools that provided broader scientific insights. In the process, we demonstrated the importance of prompts for foundation models to successfully delegate and finish tasks, the importance of benchmark evaluation, and usability of cloud infrastructure for such a deployment. We expect the copilot to be a blueprint for other scientific domains. 

Some of the limitations of the work are listed as follows: 

{\bf Use of Langraph:} While effective for lightweight task routing, LangGraph's structural constraints limit more dynamic agentic interactions. Future work will explore customized orchestration back-ends that can support richer control flow and tighter integration with scientific compute environments. 

{\bf Prompt Engineering:} A further limitation arises from the supervisor agent's reliance on semantic similarity and lexical overlap for routing decisions, which proved insufficient in disambiguating functionally distinct but linguistically overlapping tasks (e.g. data analysis vs. image segmentation analysis). To address this, future work will focus on refining the supervisor's prompt to include disambiguating examples and clearer role definitions.

{\bf Session Memory:} Our system currently maintains no persistent session memory. All intermediate results, agent decisions, and user interactions exist only in the active runtime and are lost if the user refreshes the interface or disconnects. This limits the system's ability to support iterative workflows, follow-up queries, or context-aware cumulative reasoning. In future work, we plan to implement persistent session storage that logs user history and agent interactions. The stored context can be retrieved and injected into subsequent LLM prompts, enabling more coherent multi-turn interactions. 

{\bf Human-in-the-loop:} Another promising direction is the integration of human-in-the-loop steering within the workflow execution process. Although the current system performs tasks autonomously once initiated, many scientific workflows benefit from intermediate checkpoint or expert validation. Incorporating capability that allows subject-matter experts to approve, revise, or redirect agent decision would increase the system's reliability and also better align with real-world scientific practices.

{\bf User-experience:} Streaming chatbot responses over HTTP uses a streaming content-type over a longer-lived connection which is not the most idiomatic way to design REST APIs. We ran into integration issues and error masking in multiple places and did not have time to fully debug and support streaming responses after splitting the application into front-end and back-end components. This will be an important consideration in future work as look to improve the user experience. 

\section{Acknowledgments}
\label{sec:acknowledgment}
We would like to thank domain scientists Jan Strube, Natalie Isenberg, and Dongjae Shin for developing the scientific capabilities that are utilized in this framework. This work was supported by the U.S. Department of Energy Office of Science and was done at Pacific Northwest National Laboratory, a multi-program national laboratory operated by Battelle Memorial Institute for the U.S. Department of Energy under Contract DE-AC05–76RLO1830. This paper has been cleared by PNNL for public release as PNNL-SA-214505.

\bibliography{paper}
\bibliographystyle{ACM-Reference-Format}

\section*{Appendix}
\label{appexdix}
\section{Prompts to Generate Test Cases}
\label{sec:appx_test_case}

\begin{lstlisting}[label=lst:test-hyp, language=Python, caption=Agent prompt to generate the test queries for the hypothesis generator agent.]
prompt = """
You are an agent capable of formulating and structuring a scientific research plan. Use the provided hypothesis_generator tool to generate a hypothesis and research plan. The tool generates scientific hypothesis based on the user input. If the tool returns a plan, use it verbatim without any rephrasing or paraphrasing.

# Steps
1. Access the provided tool to generate a scientific hypothesis and research plan.
2. If the tool returns a hypothesis, include it as-is in the output without modification.
3. If the tool fails to generate a hypothesis or does not provide results, proceed to construct a scientific research 
hypothesis based on the input parameters or topic provided by the user.

# Output Format
- If the tool provides a hypothesis and research plan:
```
The hypothesis generated by the tool: [Insert hypothesis and research plan exactly as returned by the tool].
```
- If a manually generated hypothesis is needed:
```
Based on the input parameters, the formulated hypothesis is: [Insert manually constructed hypothesis here].
```

# Notes    
- Ensure the hypothesis follows scientific reasoning and structure if constructed manually.
- Clearly differentiate between directly provided outputs from the tool and manually formulated hypotheses.
- Do not alter or summarize the hypothesis generated by the tool. Use it verbatim.
"""
\end{lstlisting}

\begin{lstlisting}[label=lst:test-data, language=Python, caption=Agent prompt to generate the test queries for the data analysis agent.]
prompt = """
You are a specialized Data Analysis Agent. Your primary responsibility is to analyze data using the designated tool and interpret results only when necessary.

# Guidelines
1. **Primary Task: Verbatim Output from the Tool**
- If the designated tool produces an analysis, **output the tool's result exactly as provided**. Do not modify, reinterpret, summarize, or add to the result in any way.

2. **Fallback Task: Perform the Analysis**
- If the designated tool fails to produce any analysis or its output is empty, you must perform the analysis yourself.
- Use appropriate reasoning and clearly communicate your conclusions, supported by any required justification.

3. **Neutral Tone**
- Maintain a professional and neutral tone in your output. Avoid emotional language or subjective interpretations.

4. **Conclusion Placement**
- For all fallback analyses, clearly demonstrate your reasoning steps **before** stating your final conclusions.

# Steps
When the tool returns valid analysis:
1. Receive and evaluate the tool's output for validity.
2. If valid, **directly output the tool's result verbatim** without further processing.
3. If you recieve a URL in a tool result, do not modify or summarize it. Do not truncate, wrap, shorten or split 
it. Preserve the entire string exactly.

When the tool fails to return analysis:
1. Identify the absence or inadequacy of the tool's result.
2. Analyze the data manually using domain knowledge and logical reasoning.
3. Clearly document reasoning steps in a methodical order.
4. State your final conclusion after reasoning.

# Output Format
- **When Tool Produces Results:** Pure text verbatim, exactly as the tool produces, without explanations or formatting changes.
- **When Tool Fails to Produce Results:** Provide the following structured response:
    - **Reasoning:** [Step-by-step explanation of your analysis process.]
    - **Conclusion:** [Final results of the analysis.]
"""
\end{lstlisting}

\section{Addendum Prompt for Evaluation}
\label{sec:appx_eval_prompt}

\begin{lstlisting}[label=lst:prompt-add, language=Python, caption=Prompt added to the test cases before execution.]
addendum_prompt = """
In addition, please also mention:
1. The list of sub-agents utilized to solve this query.
2. The specific tools used by those sub-agents to answer the question.

Only name the sub-agents and the tools that were actually utilized
"""
\end{lstlisting}









\end{document}